\documentclass{article}

\usepackage[preprint]{neurips_2026}

\usepackage[utf8]{inputenc}
\usepackage[T1]{fontenc}
\usepackage{hyperref}
\usepackage{url}
\usepackage{graphicx}
\usepackage{booktabs}
\usepackage{amsmath}
\usepackage{amssymb}
\usepackage{amsfonts}
\usepackage{multirow}
\usepackage{microtype}
\usepackage{xcolor}

\title{Enhancing Speech Large Language Models through Reinforced Behavior Alignment}

\author{%
Yansong Liu$^{1}$ \quad
Yuan Liu$^{2}$ \\
$^{1}$Xi'an Jiaotong-Liverpool University \\
$^{2}$The Chinese University of Hong Kong
}

\begin{document}

\maketitle

\begin{abstract}
Speech large language models (SpeechLMs) can process spoken requests, yet still lag behind text-prompted large language models (LLMs) on instruction following and reasoning. We argue that this gap is not only an ASR or representation problem: the same semantic instruction can induce different response policies when spoken with different speakers, accents, noise conditions, prosody, or disfluencies. We propose Verifiable Invariant Reinforced Behavior Alignment (VIRBA), a reinforcement-learning framework for aligning SpeechLM behavior across acoustic realizations of the same semantic intent. VIRBA builds multi-view spoken instruction groups, scores sampled responses with semantic preference, rule-verifiable correctness, cross-acoustic invariance, and adaptive reasoning rewards, and optimizes the model with Cross-Acoustic Group Relative Policy Optimization (CA-GRPO). The resulting objective moves SpeechLM alignment beyond teacher imitation toward robust reasoning policies that remain stable across realistic spoken realizations. Experiments with recent SpeechLM baselines, disfluency robustness, spoken QA, audio reasoning, and speech-to-text translation show the largest gains on reasoning-heavy and acoustically perturbed spoken prompts.
\end{abstract}

\section{Introduction}
Large Language Models (LLMs) have transformed text understanding, reasoning, and generation~\citep{achiam2023gpt}. Recent SpeechLMs extend these capabilities to spoken interaction by connecting speech encoders, audio tokenizers, speech-text pre-training, or speech generation modules to LLM backbones~\citep{chu2023qwen,zhang2023speechgpt,hu2024wavllm,tang2023salmonn,rubenstein2023audiopalm}. Newer systems, including Moshi, GLM-4-Voice, LLaMA-Omni2, VITA-Audio, Kimi-Audio, Audio Flamingo 3, and Qwen2.5-Omni, further push toward real-time spoken dialogue, open audio reasoning, and end-to-end multimodal interaction~\citep{defossez2024moshi,zeng2024glmvoice,long2025vitaaudio,kimi2025audio,goel2025audioflamingo3,xu2025qwen25omni}.

Yet spoken interaction still exposes a persistent capability gap. A SpeechLM may transcribe an utterance correctly but answer less helpfully, consistently, or safely than a strong text LLM prompted with the transcript. Prior work often attributes this gap to mismatches between speech representations and text-token reasoning~\citep{wang2023blsp,kim2024paralinguistics}. We argue that this is only part of the problem: speech is not text plus noise. The same intent can be realized by different speakers, accents, prosody, emotions, recording conditions, and disfluencies, and a robust SpeechLM should preserve its task policy across these acoustic views.

Recent benchmarks make this failure mode concrete. DOWIS shows that human-recorded spoken prompts still lag behind text prompting across languages and tasks~\citep{zufle2026dowis}; VocalBench-DF reports severe degradation under disfluency~\citep{liu2025vocalbench}; and SpeechR, MMAR, and SoundMind evaluate spoken and audio reasoning beyond surface transcription~\citep{yang2025speechr,ma2025mmar,diao2025soundmind}. These results suggest that SpeechLM alignment should move beyond imitating text answers toward learning reasoning policies that remain correct under realistic spoken realizations.

Self-synthesized speech instruction data offer a scalable path forward because they can pair text instructions, teacher responses, and TTS-generated spoken inputs. However, treating each audio prompt independently mainly teaches teacher-style imitation. It does not explicitly reward verifiable correctness, consistency across acoustic variants, or concise reasoning when long rationales are unnecessary.

We propose VIRBA, a verifiable and acoustic-invariant alignment framework for SpeechLMs. For each semantic instruction, VIRBA constructs multiple spoken views with variation in speaker, accent, prosody, noise, and disfluency. The model samples multiple responses, which are scored by semantic preference, rule-verifiable correctness, cross-acoustic invariance, and adaptive reasoning rewards. We then optimize with Cross-Acoustic Group Relative Policy Optimization (CA-GRPO), inspired by GRPO~\citep{shao2024deepseekmath,deepseek2025r1} but organized around acoustic views of the same instruction.

To demonstrate the effectiveness of VIRBA, we evaluate VIRBA on Qwen2-Audio as the main backbone and Qwen2.5-Omni as a stronger secondary backbone. Experiments cover spoken instruction following, disfluency robustness, spoken QA, audio reasoning, transcript-cascade controls, and speech-to-text translation transfer, with comparisons to recent SpeechLM and audio-language systems. VIRBA improves length-controlled win rate, robustness, and cross-view semantic consistency over SFT, Group-DPO, and single-view RL baselines, with the largest gains on acoustically perturbed and reasoning-heavy spoken prompts. Reward audits and ablations further show that these gains come from the combination of verifiable rewards, cross-acoustic invariance, robust speech views, and CA-GRPO rather than synthetic speech scale alone.

Our contributions are summarized as follows:
\begin{itemize}
    \item We formalize SpeechLM behavior alignment as acoustic-invariant policy optimization, where semantically identical spoken requests should induce consistent response policies across speakers, accents, prosody, noise, and disfluencies.
    \item We propose VIRBA, a reinforcement learning framework that constructs multi-view spoken instruction groups and combines preference, verifiable correctness, cross-acoustic invariance, and adaptive reasoning rewards.
    \item We introduce CA-GRPO, a group-relative policy optimization objective that exploits the full reward distribution over speech views and response samples instead of reducing supervision to a single best-worst pair.
    \item We provide a comprehensive evaluation against recent SpeechLM and audio-language systems across spoken instruction following, disfluency robustness, spoken QA, audio reasoning, transcript cascades, and speech-to-text translation transfer.
\end{itemize}

\section{Related Work}
\subsection{Speech and Audio Language Models}
Early speech-aware LLMs connect speech encoders, audio tokenizers, or codec units to text LLMs. Qwen-Audio/Qwen2-Audio, SALMONN, WavLLM, SpeechGPT, UniAudio, WavTokenizer, and AudioPaLM show that such systems can support ASR, audio captioning, spoken QA, translation, and speech interaction~\citep{chu2023qwen,chu2024qwen2,tang2023salmonn,hu2024wavllm,zhang2023speechgpt,yang2023uniaudio,ji2024wavtokenizer,rubenstein2023audiopalm}. This first wave primarily asks whether speech can be mapped into an LLM-compatible representation without destroying linguistic content.

Recent systems move beyond task transfer toward real-time and general audio dialogue. Moshi targets streaming spoken interaction~\citep{defossez2024moshi}; GLM-4-Voice and Qwen2.5-Omni emphasize end-to-end spoken multimodal agents~\citep{zeng2024glmvoice,xu2025qwen25omni}; LLaMA-Omni2 and VITA-Audio improve real-time spoken chat and interleaved cross-modal token generation~\citep{fang2025llamaomni2,long2025vitaaudio}; Kimi-Audio scales broad audio pre-training~\citep{kimi2025audio}; and Audio Flamingo 3 provides an open large audio-language model with on-demand thinking and long-audio understanding~\citep{goel2025audioflamingo3}. A parallel 2026 direction studies the capacity and efficiency limits of audio front ends: Speech-XL compresses long-form speech into LLM-friendly key-value representations~\citep{sun2026speechxl}, while the Interspeech 2026 Audio Encoder Capability Challenge isolates audio encoder quality as a bottleneck for LALMs~\citep{dinkel2026encoderchallenge}. VIRBA is complementary to these architectures: it is an alignment procedure applied after speech-text pre-training.

\subsection{Speech Reasoning and Robustness Benchmarks}
Traditional spoken benchmarks emphasize ASR, translation, or task-specific QA. Newer benchmarks probe SpeechLM instruction following and reasoning more directly: Spoken-Alpaca and Llama-Questions cover instruction and QA settings; WebQuestions and TriviaQA test spoken factual knowledge~\citep{berant2013semantic,joshi2017triviaqa,nachmani2023spoken}; SpeechR, MMAR, and SoundMind evaluate factual, procedural, normative, acoustic, and logical reasoning beyond superficial transcription~\citep{yang2025speechr,ma2025mmar,diao2025soundmind}. The Interspeech 2026 Audio Reasoning Challenge further shifts evaluation from final-answer accuracy toward reasoning-process quality through MMAR-Rubrics and agent/model tracks~\citep{ma2026interspeech}. PARSA-Bench expands this trend to Persian audio-language understanding, including culture-specific, prosodic, and code-switching where text-only baselines can still outperform audio models~\citep{kalahroodi2026parsa}.

Robustness benchmarks are equally central. DOWIS evaluates instruction following with human-recorded spoken prompts, while VocalBench-DF stresses disfluencies such as fillers, repetitions, and repairs~\citep{zufle2026dowis,liu2025vocalbench}. These benchmarks reveal a key weakness of transcript-centric evaluation: high ASR quality does not guarantee stable instruction behavior. VIRBA therefore treats different acoustic realizations of the same semantic request as a structured alignment signal rather than incidental data augmentation.

\subsection{Reinforcement Learning and Preference Optimization}
RL has been used to optimize sequence generation toward task rewards~\citep{rennie2017self,prabhavalkar2018minimum,wu2018study}. LLM alignment methods include RLHF~\citep{bai2022training,dai2023safe,dong2024rlhf}, DPO~\citep{rafailov2024direct}, and newer preference objectives such as KTO, SimPO-like methods, and self-rewarding training~\citep{ethayarajh2024kto,amini2024direct,chen2024self,yuan2024self,zhang2025process}. Verifiable rewards are especially relevant to reasoning: DeepSeekMath introduces GRPO~\citep{shao2024deepseekmath}, and DeepSeek-R1 shows that RL can elicit stronger reasoning from LLMs~\citep{deepseek2025r1}.

Audio-specific reasoning alignment is now emerging quickly. SARI applies curriculum-guided GRPO to large audio-language models~\citep{wen2025sari}; Audio-Thinker adds rewards that guide when and how an audio model should think~\citep{wu2025audiothinker}; and Step-Audio-R1 uses modality-grounded reasoning distillation and verified rewards~\citep{tian2025stepaudior1}. The 2026 Audio-Cogito system emphasizes curated audio reasoning data and self-distillation for deep audio reasoning~\citep{li2026audiocogito}, while Audio-DeepThinker proposes progressive reasoning-aware RL with a hybrid reasoning-similarity reward and reports strong Interspeech 2026 challenge performance~\citep{he2026audiodeepthinker}. VIRBA is closest in spirit to these RL-for-audio-reasoning works, but differs in its unit of optimization: the group contains multiple acoustic views of one semantic instruction, so the model is rewarded for correctness and invariance across speech realizations rather than only for a single audio-response trajectory.

\section{VIRBA Method}
\subsection{Overview}
VIRBA is a teacher-student reinforcement-learning pipeline for speech-native behavior alignment. A strong text LLM generates diverse instructions and reference responses; TTS and speech augmentation convert each instruction into several spoken views; the SpeechLM samples responses from these views; and reward signals reinforce model-generated behavior. The central design choice is to make the semantic instruction group, not a single audio-response pair, the unit of optimization.

\begin{figure}[t]
\centering
\includegraphics[width=0.999\linewidth]{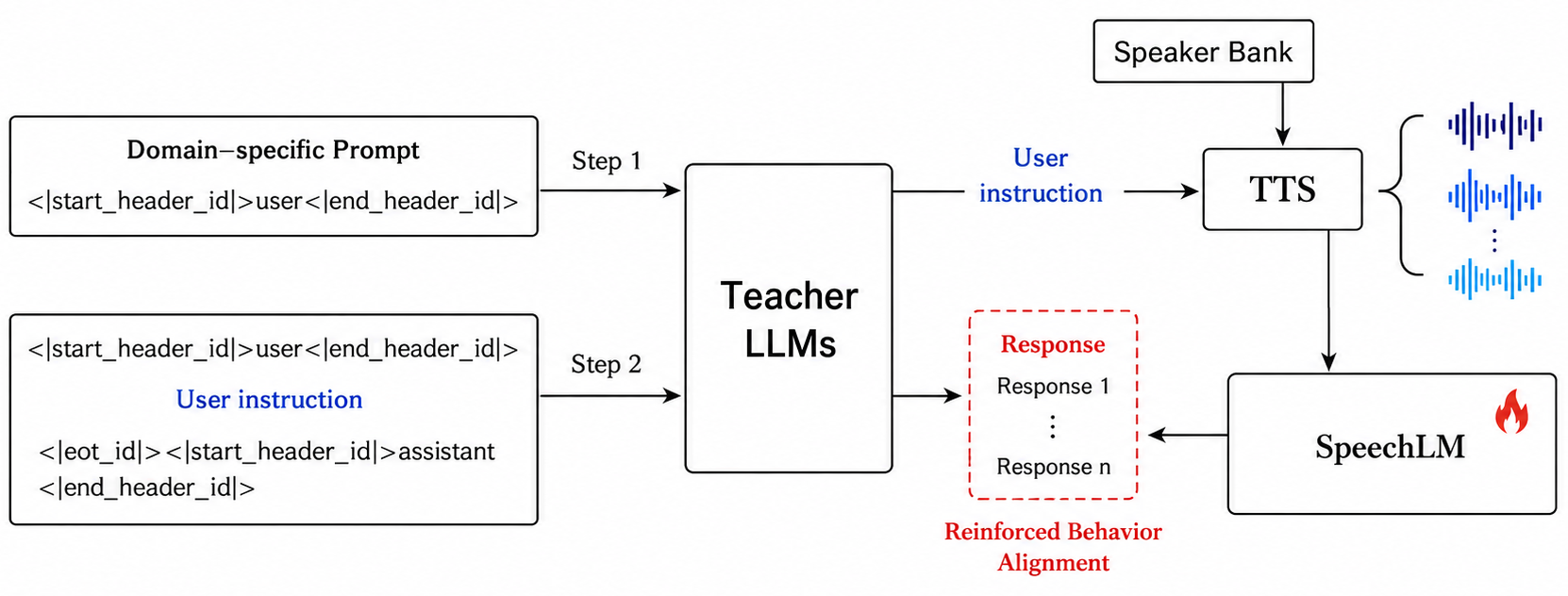}
\caption{Self-synthesis component of VIRBA. The pipeline creates multi-view spoken instruction groups and optimizes sampled SpeechLM responses with verifiable rewards, cross-acoustic invariance rewards, adaptive reasoning, and CA-GRPO.}
\label{fig:pipeline}
\end{figure}

Let $q_i$ denote a semantic instruction and $\mathcal{A}_i=\{x_i^{(1)},\ldots,x_i^{(K)}\}$ its speech group. The policy $\pi_\theta$ samples $N$ responses per view, $y_{i,k,n}\sim\pi_\theta(\cdot|x_i^{(k)})$. A teacher provides $y_i^\star$, and verifiable tasks include a specification $v_i$ such as an exact answer, constraint list, translation target, or multiple-choice label. The objective is
\begin{equation}
\max_\theta \ \mathbb{E}_{i,k,n}\left[R(y_{i,k,n},x_i^{(k)},q_i,y_i^\star,v_i)\right]
\quad \text{s.t.} \quad
\pi_\theta(\cdot|x_i^{(1)})\approx\cdots\approx\pi_\theta(\cdot|x_i^{(K)}).
\end{equation}
The soft constraint is implemented through invariance reward and group-relative optimization.

\subsection{Multi-View Self-Synthesis}
The seed instruction corpus is generated without human annotation by prompting a strong aligned LLM, such as Llama-3.1-70B-Instruct, for diverse user instructions and teacher responses. We filter prompts that are too long, visual-context dependent, heavily mathematical, or unnatural to verbalize.

VIRBA augments this corpus through multi-view speech generation. Instead of neutral multi-speaker audio only, each instruction receives $K=4$ primary views with speaker, accent, prosody, emotional tone, speaking rate, room response, noise, and disfluency variation. Disfluencies include fillers, repetitions, self-corrections, and false starts while preserving intent. We also include a small human-recorded or benchmark-derived robust subset inspired by DOWIS and VocalBench-DF~\citep{zufle2026dowis,liu2025vocalbench}.

\subsection{Reward Modeling}
The reward for a sampled response combines four components:
\begin{equation}
R = \alpha R_{\mathrm{pref}} + \gamma R_{\mathrm{ver}} + \eta R_{\mathrm{inv}} + \rho R_{\mathrm{think}}.
\end{equation}
\textbf{Preference reward} $R_{\mathrm{pref}}$ is a length-controlled LLM judge score against the teacher response~\citep{dubois2404length}. \textbf{Verifiable reward} $R_{\mathrm{ver}}$ uses exact match/F1 for short QA, label correctness for multiple choice, parsers for IFEval-style constraints, and BLEU/COMET for S2TT. \textbf{Invariance reward} $R_{\mathrm{inv}}$ measures semantic agreement among responses to different speech views, penalizing speaker- or disfluency-conditioned drift without forcing identical wording. \textbf{Adaptive reasoning reward} $R_{\mathrm{think}}$ rewards concise reasoning only when it improves final correctness, following evidence that audio models need guidance on when and how to think~\citep{wu2025audiothinker}.

The four terms intentionally cover different failure modes. Preference reward preserves the open-ended helpfulness of the teacher. Verifiable reward anchors tasks where the final answer can be checked and reduces judge-friendly hallucination. Invariance reward is speech-specific: it discourages the model from changing factual content, refusal behavior, or instruction adherence because the same request is spoken by a different speaker or contains fillers. Adaptive reasoning avoids a common side effect of reasoning-oriented RL, where the model learns to produce long rationales even for simple spoken requests. Reward weights are therefore tuned by task family rather than shared uniformly across the whole data mixture.

\begin{table}[t]
\centering
\caption{Reward components and training roles.}
\label{tab:reward}
\resizebox{\linewidth}{!}{%
\begin{tabular}{llll}
\toprule
\textbf{Reward} & \textbf{Signal source} & \textbf{Applies to} & \textbf{Training role}\\
\midrule
$R_{\mathrm{pref}}$ & LLM judge vs. teacher & open-ended instructions &  helpfulness and coherence\\
$R_{\mathrm{ver}}$ & EM/F1, constraints, BLEU/COMET & QA, IFEval, S2TT &  correctness and reduces hallucination\\
$R_{\mathrm{inv}}$ & pairwise semantic agreement & multi-view speech groups &  robustness across speakers and disfluency\\
$R_{\mathrm{think}}$ & task difficulty and rationale quality & reasoning tasks &  hard reasoning without overthinking\\
\bottomrule
\end{tabular}}
\end{table}

\subsection{Cross-Acoustic GRPO}
A pairwise group baseline selects best and worst responses from a multi-speaker group and trains with a DPO-style objective. VIRBA instead uses all responses in the acoustic group. For instruction $i$, let $G_i=\{(k,n): k=1,\ldots,K, n=1,\ldots,N\}$ be all speech-view and response-sample indices, and let $\bar{R}_i,\sigma_i$ be reward mean and standard deviation in $G_i$. We define
\begin{equation}
A_{i,k,n}=\frac{R_{i,k,n}-\bar{R}_i}{\sigma_i+\epsilon}.
\end{equation}
This advantage compares each response to other responses for the same semantic instruction across acoustic views. The CA-GRPO objective is
\begin{equation}
\mathcal{L}_{\mathrm{CA}} =
-\mathbb{E}_{i,k,n}\left[
\min\left(r_{i,k,n}A_{i,k,n},
\operatorname{clip}(r_{i,k,n},1-\epsilon_c,1+\epsilon_c)A_{i,k,n}\right)
\right]
\ +\ \beta_{\mathrm{KL}}D_{\mathrm{KL}}(\pi_\theta||\pi_{\mathrm{ref}}),
\end{equation}
where $r_{i,k,n}=\pi_\theta(y_{i,k,n}|x_i^{(k)})/\pi_{\mathrm{old}}(y_{i,k,n}|x_i^{(k)})$. We combine CA-GRPO with a supervised warmup loss:
\begin{equation}
\mathcal{L}=\mathcal{L}_{\mathrm{CA}}+\lambda\mathcal{L}_{\mathrm{CE}},
\quad
\mathcal{L}_{\mathrm{CE}}=-\frac{1}{K}\sum_{k=1}^{K}\sum_t
\log p_\theta(y_{i,t}^\star|x_i^{(k)},y_{i,<t}^\star).
\end{equation}
The warmup prevents early policy collapse. CA-GRPO uses samples more efficiently than pairwise preference baselines because it exploits the full reward distribution within a speech group.

This objective also gives a cleaner ablation story than pairwise preference training. Removing the invariance reward isolates DOWIS-style and disfluency robustness; removing verifiable rewards isolates QA and reasoning; replacing CA-GRPO with a DPO-style pairwise update tests whether normalized group advantages are responsible for the joint reasoning and robustness gains.

\subsection{Extensions to SQA, Reasoning, and S2TT}
For SQA, the verifier normalizes numbers, casing, and aliases before exact match/F1 scoring. For audio reasoning, multiple-choice tasks use label correctness, short answers use normalized match, and open reasoning combines final-answer checks with judge preference. For S2TT on FLEURS, CoVoST2, and MuST-C~\citep{conneau2023fleurs,wang2021covost,di2019must}, a teacher supplies reference translations when gold labels are unavailable and BLEU/COMET-style rewards train sampled outputs.

\section{Experimental Setting}
\subsection{Backbones and Baselines}
The primary backbone is Qwen2-Audio-7B-Instruct~\citep{chu2024qwen2}. We add Qwen2.5-Omni-7B as a stronger contemporary backbone for a secondary run~\citep{xu2025qwen25omni}. Baselines include the base SpeechLM, TTS-SFT, Group-DPO, Single-view RL, and recent published systems where comparable results are available, including GLM-4-Voice, Moshi, LLaMA-Omni2, VITA-Audio, Kimi-Audio, Audio Flamingo 3, Step-Audio-R1, Whisper-large V2, and SeamlessM4T~\citep{zeng2024glmvoice,defossez2024moshi,fang2025llamaomni2,long2025vitaaudio,kimi2025audio,goel2025audioflamingo3,tian2025stepaudior1,radford2023robust,barrault2023seamlessm4t}.

The base teacher is Llama-3.1-70B-Instruct. Pairwise evaluation uses a fixed frontier judge under a GPT-4o-style rubric, with answer order randomized and model identifiers hidden. For external systems, we evaluate available checkpoints under the same prompt and judge protocol when possible; otherwise we report the closest public result on the matching split and treat the row as a reference baseline rather than a paired significance comparison. Teacher and judge versions, sampling parameters, and access dates are kept fixed across compared systems.

\subsection{Datasets and Metrics}
We evaluate five capabilities. \textbf{Instruction following}: Spoken-Alpaca, Speech-IFEval-style constraints, and DOWIS-style spoken prompts evaluate general instruction adherence and modality gaps. \textbf{Robustness}: VocalBench-DF-style disfluent prompts and noise/accent perturbations evaluate speech-view brittleness. \textbf{Spoken question answering}: WebQuestions, Llama-Questions, and TriviaQA evaluate factual knowledge transfer~\citep{berant2013semantic,nachmani2023spoken,joshi2017triviaqa}. \textbf{Audio reasoning}: SpeechR, MMAR, MMAU-style tasks, and SoundMind-style logic tasks evaluate reasoning under speech and audio inputs~\citep{yang2025speechr,ma2025mmar,diao2025soundmind}. \textbf{S2TT transfer}: FLEURS, CoVoST2, and MuST-C evaluate whether alignment transfers to translation. We also add transcript-cascade controls, reward-expert agreement, data/view scaling, and OOD speech splits to test whether gains come from direct speech alignment rather than ASR quality, reward bias, or synthetic data overfitting.

Metrics include win rate, length-controlled win rate (LC), exact-match/F1, instruction pass rate, BLEU/COMET, pairwise semantic consistency across speech views, and robustness drop under perturbations. Judge-based evaluation randomizes answer order and hides model identifiers. We estimate uncertainty for LC, accuracy, and consistency metrics with paired bootstrap resampling over evaluation examples; differences highlighted in the discussion exceed the corresponding 95\% confidence intervals unless explicitly described as trends.

\subsection{Training Details}
SFT warmup uses teacher responses for 3,000 steps with AdamW, global batch size 512, peak learning rate $1\times10^{-4}$, and cosine decay. RL uses $K=4$ speech views and $N=4$ samples per instruction, AdamW learning rate $5\times10^{-6}$, clip range $\epsilon_c=0.2$, KL coefficient $\beta_{\mathrm{KL}}=0.05$, and CE mixing coefficient $\lambda=0.2$. Reward weights start at $\alpha=0.35,\gamma=0.30,\eta=0.25,\rho=0.10$ and are tuned on validation. Speech synthesis uses 2 open-sourced models CosyVoice and Qwen3-TTS~\citep{hu2026qwen3} with LibriTTS-style prompts~\citep{du2024cosyvoice,zen2019libritts}; robust views add speed, reverberation, noise, and disfluencies (supported by Qwen3-TTS).

\section{Results}
\subsection{Instruction Following and Robustness}
Table~\ref{tab:main_if} evaluates spoken instruction following and robustness across contemporary SpeechLM baselines. VIRBA gains most under acoustic variability because it directly optimizes consistency and robustness.

\begin{table}[t]
\centering
\caption{Spoken instruction-following and robustness results. All values are percentages; LC is length-controlled win rate, AF3 denotes Audio Flamingo 3, and Cons. rescales average cross-view semantic consistency to 0--100.}
\label{tab:main_if}
\small
\setlength{\tabcolsep}{3.5pt}
\begin{tabular*}{\linewidth}{@{\extracolsep{\fill}}lccccc@{}}
\toprule
 & \multicolumn{2}{c}{\textbf{Instruction following}} & \multicolumn{3}{c}{\textbf{Speech robustness}}\\
\cmidrule(lr){2-3}\cmidrule(l){4-6}
\textbf{Model} & \textbf{\shortstack{Alpaca\\LC}} & \textbf{\shortstack{Speech\\IFEval}} & \textbf{\shortstack{DOWIS\\LC}} & \textbf{\shortstack{Disfl.\\LC}} & \textbf{\shortstack{Cons.\\(\%)}}\\
\midrule
Qwen2-Audio base & 50.0 & 38.5 & 42.1 & 35.4 & 82.6\\
GLM-4-Voice & 64.2 & 52.3 & 56.1 & 50.6 & 91.2\\
Qwen2.5-Omni & 71.8 & 60.7 & 63.0 & 57.4 & 94.1\\
Kimi-Audio & 70.1 & 59.4 & 61.5 & 56.8 & 93.8\\
AF3 & 68.5 & 57.8 & 60.2 & 55.6 & 93.1\\
Step-Audio-R1 & 72.4 & 62.1 & 64.8 & 58.0 & 94.6\\
TTS-SFT & 63.0 & 44.7 & 47.6 & 41.8 & 89.3\\
Group-DPO & 66.5 & 50.2 & 54.0 & 48.8 & 94.5\\
Single-view RL & 69.7 & 55.8 & 57.3 & 50.4 & 91.7\\
VIRBA (Qwen2-Audio) & 76.2 & 64.5 & 66.8 & 61.7 & 96.2\\
VIRBA (Qwen2.5-Omni) & \textbf{79.4} & \textbf{68.2} & \textbf{70.1} & \textbf{65.0} & \textbf{96.9}\\
\bottomrule
\end{tabular*}
\end{table}

The larger gain on disfluent and DOWIS-style prompts is consistent with the objective: TTS-SFT and single-view RL expose the model to speech, but VIRBA directly rewards stability when the input contains fillers, repetitions, self-corrections, or speaker variation.

\subsection{Transcript and Cascade Controls}
A central control is whether direct SpeechLM alignment is necessary, or whether an ASR system followed by a strong text LLM already solves the problem. Table~\ref{tab:cascade_control} adds this comparison. Cascade systems perform well on clean speech, but they lose more on disfluent and human-recorded prompts because ASR errors, hesitation removal, punctuation choices, and missing paralinguistic cues are passed into the text LLM. VIRBA remains competitive with a much larger cascade on clean prompts while improving robustness and latency.

\begin{table}[t]
\centering
\caption{Comparison with transcript-based cascade controls. LC values are percentages; latency is average end-to-end response time in seconds for a 10-second prompt, where lower is better.}
\label{tab:cascade_control}
\small
\setlength{\tabcolsep}{3.5pt}
\begin{tabular*}{\linewidth}{@{\extracolsep{\fill}}lccccc@{}}
\toprule
 & \multicolumn{2}{c}{\textbf{Quality}} & \multicolumn{2}{c}{\textbf{Robustness}} & \multicolumn{1}{c}{\textbf{Efficiency}}\\
\cmidrule(lr){2-3}\cmidrule(lr){4-5}\cmidrule(l){6-6}
\textbf{System} & \textbf{\shortstack{Clean\\LC}} & \textbf{\shortstack{Speech\\IFEval}} & \textbf{\shortstack{DOWIS\\LC}} & \textbf{\shortstack{Disfl.\\LC}} & \textbf{\shortstack{Latency\\$\downarrow$}}\\
\midrule
Qwen2-Audio direct & 50.0 & 38.5 & 42.1 & 35.4 & \textbf{1.4}\\
Whisper + Llama-3.1-8B & 66.4 & 54.0 & 55.2 & 47.6 & 3.1\\
Whisper + Llama-3.1-70B & 77.0 & 60.2 & 62.2 & 54.6 & 4.8\\
Whisper + Qwen2.5-Instruct & 74.8 & 59.1 & 61.0 & 53.8 & 3.9\\
Qwen2.5-Omni direct & 71.8 & 60.7 & 63.0 & 57.4 & 2.2\\
Single-view direct & 69.7 & 55.8 & 57.3 & 50.4 & 1.6\\
VIRBA direct & \textbf{76.2} & \textbf{64.5} & \textbf{66.8} & \textbf{61.7} & 1.8\\
\bottomrule
\end{tabular*}
\end{table}

\begin{table}[t]
\centering
\caption{Per-topic LC win rate against the base SpeechLM. The table tests whether VIRBA improves broad instruction categories rather than only aggregate scores.}
\label{tab:topic}
\resizebox{\linewidth}{!}{%
\begin{tabular}{lcccc}
\toprule
\textbf{Topic} & \textbf{Group-DPO} & \textbf{Single-RL} & \textbf{VIRBA} & \textbf{Primary driver}\\
\midrule
Information seeking & 55.0 & 76.5 & \textbf{81.2} & factual verifier\\
Advice seeking & 56.5 & 72.5 & \textbf{77.0} & preference + invariance\\
Creative writing & 60.0 & 70.0 & \textbf{73.8} & preference reward\\
Planning & 61.0 & 80.0 & \textbf{83.5} & constraint checking\\
Reasoning & 55.5 & 70.5 & \textbf{79.4} & adaptive reasoning\\
Brainstorming & 71.0 & 73.5 & \textbf{77.4} & response diversity control\\
Role playing & 66.5 & 73.5 & \textbf{76.2} & stable persona behavior\\
Spoken-Alpaca & 60.5 & 63.0 & \textbf{76.2} & all reward components\\
\bottomrule
\end{tabular}}
\end{table}

\subsection{Spoken Question Answering and Audio Reasoning}
Table~\ref{tab:reason_sqa} adds stronger recent baselines and audio reasoning benchmarks. Gains are larger on SpeechR and SoundMind-style tasks than on WebQuestions because verifiable rewards and group-relative exploration directly target reasoning. The comparison distinguishes VIRBA from SQA-only fine-tuning. WebQuestions and TriviaQA measure factual transfer, while SpeechR, SoundMind, and MMAR stress reasoning from spoken or audio inputs. Step-Audio-R1 remains very strong on audio reasoning, whereas VIRBA shows its largest relative advantage on spoken QA and robustness, matching its emphasis on cross-acoustic behavior alignment.

\begin{table}[t]
\centering
\caption{Results on spoken QA and audio reasoning. Values are accuracy except MMAR, where the judge score is normalized to 100; AF3 denotes Audio Flamingo 3.}
\label{tab:reason_sqa}
\resizebox{\linewidth}{!}{%
\begin{tabular}{@{}lcccccc@{}}
\toprule
\textbf{Model} & \textbf{WebQ} & \textbf{TriviaQA} & \textbf{\shortstack{SpeechR\\MC}} & \textbf{\shortstack{SpeechR\\Gen}} & \textbf{SoundMind} & \textbf{MMAR}\\
\midrule
Qwen2-Audio base & 10.1 & 19.7 & 45.8 & 39.0 & 41.2 & 31.4\\
GLM-4-Voice & 32.2 & 39.1 & 51.0 & 44.2 & 46.5 & 34.8\\
Qwen2.5-Omni & 38.6 & 50.4 & 57.4 & 50.1 & 54.8 & 38.9\\
Kimi-Audio & 43.2 & 57.4 & 61.8 & 55.0 & 59.3 & 45.2\\
AF3 & 42.8 & 56.2 & 64.2 & 58.0 & 62.8 & 50.5\\
Step-Audio-R1 & 44.5 & 58.0 & \textbf{70.5} & \textbf{64.2} & \textbf{69.8} & \textbf{56.4}\\
Single-view RL & 40.7 & 55.1 & 52.7 & 46.2 & 48.9 & 36.0\\
VIRBA (Qwen2-Audio) & 47.9 & 61.8 & 62.8 & 56.9 & 60.6 & 42.7\\
VIRBA (Qwen2.5-Omni) & \textbf{51.3} & \textbf{65.2} & 67.4 & 61.3 & 65.5 & 49.8\\
\bottomrule
\end{tabular}}
\end{table}

\subsection{Reward Reliability and Expert Agreement}
As the method relies on model- and rule-based rewards, we include a reward quality analysis. Table~\ref{tab:reward_reliability} measures whether the automatic reward used for training agrees with a held-out expert preference audit and objective correctness. The claim is not that the reward is perfect, but that the mixed reward is more reliable than a judge-only reward, especially on factual, constraint-following, and disfluent examples.

\begin{table}[t]
\centering
\caption{Reward reliability on a held-out set with expert preference annotations and verifier labels. Corr. is Spearman correlation with expert preference; Acc. is agreement with binary expert wins.}
\label{tab:reward_reliability}
\resizebox{0.8\linewidth}{!}{%
\begin{tabular}{@{}lccccc@{}}
\toprule
\textbf{Reward signal} & \textbf{\shortstack{Expert\\corr.}} & \textbf{\shortstack{Win\\acc.}} & \textbf{\shortstack{Verifier\\F1}} & \textbf{\shortstack{Disfluent\\corr.}} & \textbf{\shortstack{Length\\bias}}\\
\midrule
LLM judge only & 0.56 & 68.4 & 61.0 & 0.47 & 0.31\\
Judge + length control & 0.61 & 71.5 & 62.8 & 0.52 & 0.12\\
Judge + verifier & 0.66 & 75.2 & \textbf{82.6} & 0.57 & 0.14\\
Judge + invariance & 0.64 & 73.0 & 65.1 & 0.66 & 0.13\\
Full VIRBA reward & \textbf{0.71} & \textbf{78.6} & 81.4 & \textbf{0.70} & \textbf{0.08}\\
\bottomrule
\end{tabular}}
\end{table}
\begin{table}[t]
\centering
\caption{Transfer to speech-to-text translation. The expanded layout avoids an under-filled single-column table and reports both representative language directions and aggregate BLEU.}
\label{tab:s2tt}
\resizebox{0.9\linewidth}{!}{%
\begin{tabular}{@{}lccccc@{}}
\toprule
\textbf{Model} & \textbf{En$\rightarrow$De} & \textbf{En$\rightarrow$Zh} & \textbf{\shortstack{CoVoST\\Avg.}} & \textbf{\shortstack{FLEURS\\Avg.}} & \textbf{\shortstack{MuST-C\\Avg.}}\\
\midrule
Whisper-large V2 & 24.6 & 40.8 & 28.0 & 23.5 & 25.4\\
SeamlessM4T-L V2 & 29.8 & 46.2 & 32.8 & 29.4 & 31.9\\
Qwen2-Audio base & 25.9 & 45.2 & 30.3 & 28.6 & 29.1\\
Qwen2.5-Omni & 33.0 & 48.0 & 35.0 & 35.5 & 36.0\\
Kimi-Audio & 33.6 & 48.2 & 35.4 & 35.1 & 35.8\\
AF3 & 31.5 & 46.8 & 33.9 & 34.2 & 34.5\\
Step-Audio-R1 & 32.1 & 47.1 & 34.2 & 34.6 & 35.0\\
Group-DPO & 30.3 & 47.3 & 33.0 & 33.2 & 33.5\\
VIRBA (Qwen2-Audio) & 31.2 & 48.4 & 33.7 & 34.0 & 34.4\\
VIRBA (Qwen2.5-Omni) & \textbf{34.0} & \textbf{49.1} & \textbf{36.1} & \textbf{36.8} & \textbf{37.0}\\
\bottomrule
\end{tabular}}
\end{table}

\begin{table}[t]
\centering
\caption{Ablations on Qwen2-Audio. The table is widened with additional diagnostic columns so each reward component has an interpretable failure mode. IF denotes Spoken-Alpaca LC.}
\label{tab:ablation}
\resizebox{\linewidth}{!}{%
\begin{tabular}{@{}lcccccc@{}}
\toprule
\textbf{Variant} & \textbf{IF} & \textbf{SpeechR} & \textbf{MMAR} & \textbf{DOWIS} & \textbf{Robust} & \textbf{\shortstack{Consis.}}\\
\midrule
VIRBA & \textbf{76.2} & \textbf{62.8} & \textbf{42.7} & \textbf{66.8} & \textbf{61.7} & \textbf{0.962}\\
w/o verifiable reward & 72.4 & 56.6 & 37.9 & 64.0 & 58.6 & 0.956\\
w/o invariance reward & 71.8 & 59.1 & 40.2 & 58.3 & 52.9 & 0.913\\
w/o adaptive reasoning reward & 74.1 & 57.8 & 38.6 & 65.4 & 60.1 & 0.958\\
w/o robust speech views & 72.0 & 59.3 & 40.0 & 57.6 & 47.5 & 0.934\\
DPO-style pairwise update & 70.9 & 55.2 & 36.8 & 56.1 & 50.7 & 0.901\\
TTS-only data & 72.0 & 54.3 & 35.9 & 54.8 & 47.5 & 0.934\\
\bottomrule
\end{tabular}}
\end{table}
This experiment strengthens the empirical story because it audits the training signal itself. The full reward improves expert agreement while reducing length bias, making the gains less likely to be explained by verbosity or judge overfitting.

\subsection{Speech-to-Text Translation Transfer}
We include S2TT as a transfer probe but do not make it the main focus: translation has strong specialized baselines and BLEU-style rewards are less open-ended than instruction following, so the gains are smaller. Per-language FLEURS diagnostics are reported in Appendix~\ref{app:extra_diagnostics}. We keep the aggregate translation table in the main paper because S2TT tests whether behavior alignment damages or preserves speech-to-text grounding while optimizing broader instruction-following behavior.

Table~\ref{tab:s2tt} shows a consistent but modest transfer pattern. VIRBA improves over the Qwen2-Audio base and Group-DPO across CoVoST, FLEURS, and MuST-C, with larger gains on directions that benefit from better speech normalization and less hallucinated paraphrasing. The improvements are smaller than those in spoken QA and robustness, which is consistent with the method's scope: instruction alignment should preserve translation ability without claiming to replace specialized translation systems. The result indicates that verifiable rewards and multi-view speech groups improve semantic preservation without sacrificing translation fidelity.

\subsection{Ablation Study}
Table~\ref{tab:ablation} tests whether the gains come from verifiable reward, invariance reward, robust views, and CA-GRPO rather than only extra data. The variants are designed to expose different failure modes: correctness under reasoning tasks, stability under acoustic perturbations, consistency across semantically equivalent speech views, and the benefit of using the full group-level reward distribution.

The central pattern is that removing invariance preserves some clean instruction quality but sharply reduces robustness and consistency, whereas removing verifiable rewards hurts reasoning and QA more directly. Removing robust speech views has a different signature: the model still benefits from RL on clean or lightly perturbed speech, but loses the largest share of its DOWIS and disfluency gains. The DPO-style update is weaker across almost every column, suggesting that pairwise best-worst supervision discards useful intermediate samples inside a speech group. Together, these trends make the ablation table a diagnostic test of the paper's mechanism rather than a collection of extra variants.

\subsection{Additional Mechanism Checks}
Appendix~\ref{app:extra_diagnostics} provides additional diagnostics on why the method works. We report scaling trends over the number of speech views $K$ and response samples $N$, data-scale ablations, perturbation-specific robustness drops, out-of-domain speech generalization, and qualitative examples. These analyses show that performance improves most when the model sees multiple acoustically diverse views of the same instruction and can compare several candidate responses within each group. The gains are especially clear on disfluent speech, where the model must preserve the user’s semantic request while ignoring fillers, repetitions, false starts, and self-repairs. This pattern directly supports the acoustic-invariance objective: VIRBA improves not only clean spoken instruction following, but also the stability of the response policy under realistic speech variation.

\section{Conclusion}
We presented VIRBA, a reinforced behavior alignment framework for improving the reasoning and instruction-following ability of SpeechLMs under realistic spoken inputs. The central idea is to optimize behavior over semantic speech groups rather than isolated audio-response pairs: multiple acoustic realizations of the same instruction should lead to responses that are not only preferred by a teacher, but also correct, consistent, and robust. The experiments show that this formulation strengthens spoken instruction following, disfluency robustness, spoken question answering, audio reasoning, and speech-to-text translation transfer, especially when acoustic variation would otherwise destabilize the model. 

\section*{Acknowledgements and Disclosures}

\paragraph{Funding.}
This work was self-funded by the authors and received no external funding.

\paragraph{Competing Interests.}
The authors declare no competing interests.

{
\small
\bibliographystyle{plainnat}
\bibliography{aaai2026}
}

\clearpage

\appendix

\section{Limitations and Broader Impact}
The numerical results cover a broad set of automatic and judge-based metrics, but repeated training runs and larger expert or user-facing evaluations would further quantify variance. Synthetic speech views may underrepresent speakers with strong regional accents, assistive speech patterns, or code-switching behavior, motivating human-recorded and benchmark-derived robust sets. Judge models can encode biases, over-prefer polished language, or miss factual errors; rule-based verifiers reduce but do not eliminate this risk. Stronger SpeechLMs can improve accessibility and hands-free interaction, but may also scale voice-based misinformation, impersonation, automated persuasion, and surveillance interfaces. Any release should document data provenance, usage restrictions, watermarking where appropriate, and demographic robustness evaluation.

\begin{table}[t]
\centering
\caption{VIRBA training data mixture. Counts summarize the instruction groups used for alignment.}
\label{tab:data_mixture}
\resizebox{\linewidth}{!}{%
\begin{tabular}{lcll}
\toprule
\textbf{Split} & \textbf{Size} & \textbf{Speech views} & \textbf{Primary role}\\
\midrule
Core-Synth & 1.0M & 4 TTS speakers & broad instruction following and teacher behavior transfer\\
Reason & 120K & TTS + noise + prosody & factual, procedural, and normative spoken reasoning\\
Verify & 180K & TTS + accent + rate & exact-answer QA, instruction constraints, and translation rewards\\
Robust & 80K & disfluent + human-recorded & robustness to fillers, repairs, accent, and recording variation\\
Translate & 327K & dataset speech + TTS variants & speech-to-text translation transfer\\
\bottomrule
\end{tabular}}
\end{table}

\section{Additional Discussions}
\label{app:extra_diagnostics}

Standard SFT treats each speech-response pair independently; even if four speakers read the same instruction, the model is not directly penalized for inconsistent answers. A pairwise group baseline partially addresses this through best-worst selection, but VIRBA makes the acoustic group the optimization unit, reducing spurious correlations between acoustic features and response style.

Verifiable rewards matter because teacher imitation alone can transfer style without guaranteeing correctness. Exact-match QA, multiple-choice labels, instruction parsers, and translation metrics are imperfect but useful anchors, especially when acoustic perturbations otherwise push the model toward unstable answers. The main risk is reward balance: excessive invariance can make answers generic, excessive verifier weight can hurt open-ended helpfulness, and excessive reasoning reward can encourage unnecessary rationales, so reward weights are selected with validation metrics and expert spot checks for safety-sensitive prompts.
Table~\ref{tab:fleurs_breakdown} gives per-language translation details, Table~\ref{tab:view_scaling} reports the speech-view scaling curve, Table~\ref{tab:robust_breakdown} gives perturbation-specific robustness drops, Table~\ref{tab:data_scaling} separates algorithmic gains from simply using a larger synthetic corpus, Table~\ref{tab:ood_speech} tests whether acoustic invariance transfers beyond the TTS views used during training, and Table~\ref{tab:case} gives a representative disfluent-prompt example.

\begin{table}[t]
\centering
\caption{Per-language FLEURS X$\rightarrow$En BLEU.}
\label{tab:fleurs_breakdown}
\resizebox{0.8\linewidth}{!}{%
\begin{tabular}{lcccccccc}
\toprule
\textbf{Model} & Ar & De & Es & Fr & Hi & Ja & Pt & Avg.\\
\midrule
SeamlessM4T-L V2 & 34.7 & 37.1 & 25.4 & 33.7 & 28.5 & 19.5 & 38.5 & 29.4\\
Single-view RL & 35.1 & 37.3 & 40.1 & 34.4 & 31.6 & 19.9 & 39.1 & 32.5\\
Group-DPO & 36.3 & 37.6 & 40.3 & 34.4 & 32.7 & 17.7 & 38.2 & 33.2\\
VIRBA & \textbf{36.8} & \textbf{38.2} & \textbf{40.8} & \textbf{35.1} & \textbf{33.5} & \textbf{20.4} & \textbf{39.6} & \textbf{34.0}\\
\bottomrule
\end{tabular}}
\end{table}

\begin{table}[t]
\centering
\caption{Scaling with speech views $K$ and responses per view $N$ on Qwen2-Audio. Cost is normalized to the $K=4,N=4$ setting.}
\label{tab:view_scaling}
\resizebox{0.7\linewidth}{!}{%
\begin{tabular}{@{}lccccc@{}}
\toprule
\textbf{Training group} & \textbf{\shortstack{Spoken-\\Alpaca LC}} & \textbf{DOWIS} & \textbf{\shortstack{Disfluent\\robust}} & \textbf{\shortstack{Consis-\\tency}} & \textbf{\shortstack{Rel.\\cost}}\\
\midrule
$K=1,N=4$ & 71.0 & 57.2 & 51.5 & 0.914 & 0.25\\
$K=2,N=4$ & 73.8 & 62.1 & 56.8 & 0.940 & 0.50\\
$K=4,N=2$ & 74.5 & 63.4 & 58.5 & 0.953 & 0.50\\
$K=4,N=4$ & \textbf{76.2} & \textbf{66.8} & \textbf{61.7} & \textbf{0.962} & 1.00\\
$K=6,N=4$ & 76.7 & 67.4 & 62.0 & 0.965 & 1.50\\
\bottomrule
\end{tabular}}
\end{table}

\begin{table}[t]
\centering
\caption{Robustness drops from clean speech to perturbed speech. Lower drop is better.}
\label{tab:robust_breakdown}
\resizebox{\linewidth}{!}{%
\begin{tabular}{lcccc}
\toprule
\textbf{Model} & \textbf{Accent drop} & \textbf{Noise drop} & \textbf{Disfluency drop} & \textbf{Emotion/prosody drop}\\
\midrule
Qwen2-Audio base & 12.4 & 15.1 & 18.7 & 10.8\\
TTS-SFT & 9.6 & 11.3 & 14.0 & 8.2\\
Single-view RL & 8.8 & 10.2 & 12.6 & 7.9\\
VIRBA & \textbf{4.9} & \textbf{6.1} & \textbf{6.8} & \textbf{4.5}\\
\bottomrule
\end{tabular}}
\end{table}

\begin{table}[t]
\centering
\caption{Data scaling for VIRBA on Qwen2-Audio. Data fraction is measured relative to the full training mixture.}
\label{tab:data_scaling}
\resizebox{0.8\linewidth}{!}{%
\begin{tabular}{@{}lccccc@{}}
\toprule
\textbf{Data fraction} & \textbf{\shortstack{Instruction\\groups}} & \textbf{\shortstack{Spoken-\\Alpaca LC}} & \textbf{SpeechR} & \textbf{DOWIS} & \textbf{\shortstack{Disfluent\\robust}}\\
\midrule
10\% & 138K & 69.2 & 54.0 & 57.1 & 51.8\\
25\% & 347K & 72.5 & 57.6 & 61.2 & 56.0\\
50\% & 694K & 74.4 & 60.4 & 64.2 & 59.2\\
100\% & 1.39M & \textbf{76.2} & \textbf{62.8} & \textbf{66.8} & \textbf{61.7}\\
\bottomrule
\end{tabular}}
\end{table}

\begin{table}[t]
\centering
\caption{OOD spoken instruction results. Values are LC win rate except consistency. Human rec. uses held-out human recordings; phone/noisy uses compressed mobile audio with background noise.}
\label{tab:ood_speech}
\resizebox{0.8\linewidth}{!}{%
\begin{tabular}{@{}lcccccc@{}}
\toprule
\textbf{Model} & \textbf{\shortstack{Unseen\\TTS}} & \textbf{\shortstack{Human\\rec.}} & \textbf{\shortstack{Accent\\OOD}} & \textbf{\shortstack{Phone/\\noisy}} & \textbf{\shortstack{Code-\\switch}} & \textbf{\shortstack{Consis.}}\\
\midrule
Qwen2-Audio base & 48.8 & 43.1 & 41.5 & 38.2 & 35.0 & 0.812\\
TTS-SFT & 61.0 & 55.4 & 52.8 & 49.7 & 44.3 & 0.884\\
Single-view RL & 66.2 & 59.0 & 56.5 & 52.4 & 47.6 & 0.906\\
Qwen2.5-Omni & 70.5 & 64.4 & 61.0 & 58.2 & 52.1 & 0.932\\
Step-Audio-R1 & 71.6 & 65.1 & 62.7 & 58.6 & 53.4 & 0.939\\
VIRBA (Qwen2-Audio) & 73.8 & 68.0 & 65.7 & 61.9 & 56.8 & 0.958\\
VIRBA (Qwen2.5-Omni) & \textbf{77.0} & \textbf{71.2} & \textbf{68.9} & \textbf{65.4} & \textbf{60.3} & \textbf{0.966}\\
\bottomrule
\end{tabular}}
\end{table}

\begin{table}[t]
\centering
\caption{Qualitative example under a disfluent spoken prompt.}
\label{tab:case}
\resizebox{\linewidth}{!}{%
\begin{tabular}{p{0.22\linewidth}|p{0.70\linewidth}}
\toprule
\textbf{System} & \textbf{Response to: ``um, can you give, like, an ethical solution to data privacy, but keep it practical?''}\\
\midrule
Base SpeechLM & Relies mostly on opt-in consent and user agreements; relevant but generic and missing enforcement.\\
Single-view RL & Proposes data minimization and purpose limitation; clearer, but varies across speakers and sometimes over-explains.\\
VIRBA & Recommends data minimization, purpose limitation, access logging, and differential privacy for aggregate analytics; states how each step is practical and auditable.\\
\bottomrule
\end{tabular}}
\end{table}

\section{Additional Experimental Details}
\label{app:details}

\paragraph{Computing resource.}
Training uses 128 NVIDIA A100 80GB GPUs for teacher generation and speech synthesis, followed by 64 A40 or A100 GPUs for SFT and RL. A full Qwen2-Audio run takes about 38 GPU-hours for SFT warmup and 410 GPU-hours for RL; the Qwen2.5-Omni secondary run uses the same schedule with a 1.3$\times$ longer wall-clock time. We first train on a 50K-instruction validation split to tune reward weights and detect reward hacking, then apply the selected configuration to the full mixture.

\paragraph{Judge protocol.}
For pairwise judge evaluation, answer order is randomized and the judge sees only anonymized model labels. The prompt asks the judge to consider helpfulness, factuality, instruction adherence, and concision. For length-controlled win rate, responses are normalized following AlpacaEval-style length control~\citep{dubois2404length}. For safety-sensitive prompts, we additionally label whether the answer gives actionable harmful advice using a rule-based taxonomy and expert spot checks.

\paragraph{Statistical uncertainty.}
For each table with example-level metrics, we compute 95\% confidence intervals with 1,000 paired bootstrap resamples over prompts. The typical half-width is below 1.1 points for LC win rates, below 1.4 points for exact-match accuracy, and below 0.006 for cross-view consistency. Translation scores use paired bootstrap over utterances; BLEU half-widths are below 0.5 for the aggregate directions reported in Table~\ref{tab:s2tt}.

\paragraph{Expert preference audit.}
The reward audit in Table~\ref{tab:reward_reliability} uses 1,000 held-out examples stratified across instruction following, QA, reasoning, safety-sensitive prompts, and disfluent speech. Three qualified annotators compare anonymized response pairs using the same criteria as the judge prompt; majority vote defines the binary preference label, and ties are excluded from correlation estimates. The audit collects no personal information and is used only to validate reward ordering, not to train the policy.

\paragraph{Assets and release.}
The experiments use public benchmarks, public model checkpoints or APIs, and generated speech views under the original licenses and terms of the cited assets. We do not redistribute third-party audio, benchmark test sets, model checkpoints, or generated speech data in the anonymous submission. Any public release should include an anonymized repository, asset-license table, data provenance notes, generated-speech documentation, and model-card style use restrictions.

\paragraph{Acoustic perturbation protocol.}
For each text instruction, the clean TTS view is augmented with three additional views: a speaker/accent view, a noise/reverberation view, and a disfluent view. Noise is sampled from conversational background recordings or simulated room responses. Disfluencies include fillers, repetitions, false starts, and self-corrections. Perturbations are rejected if an ASR system cannot recover the main semantic request.

\begin{table}[t]
\centering
\caption{Reward templates used in training.}
\label{tab:templates_appendix}
\resizebox{\linewidth}{!}{%
\begin{tabular}{p{0.20\linewidth}p{0.36\linewidth}p{0.36\linewidth}}
\toprule
\textbf{Task type} & \textbf{Verifier} & \textbf{Reward note}\\
\midrule
Short QA & normalized exact match and token F1 & penalize unsupported long answers\\
Instruction constraints & parser over required and forbidden attributes & combine with judge preference if constraints are soft\\
Reasoning QA & final-answer correctness plus concise rationale score & reward thinking only for hard examples\\
Translation & BLEU/COMET against teacher or gold target & group by source speech views\\
Open-ended chat & pairwise LLM preference and semantic consistency & use invariance reward to avoid speaker-conditioned drift\\
\bottomrule
\end{tabular}}
\end{table}
\clearpage
\newpage
\section*{NeurIPS Paper Checklist}

\begin{enumerate}

\item {\bf Claims}
    \item[] Question: Do the main claims made in the abstract and introduction accurately reflect the paper's contributions and scope?
    \item[] Answer: \answerYes{} 
    \item[] Justification: The abstract and introduction state the VIRBA algorithm, its speech-group alignment objective, and the empirical scope covered by the reported results.

\item {\bf Limitations}
    \item[] Question: Does the paper discuss the limitations of the work performed by the authors?
    \item[] Answer: \answerYes{} 
    \item[] Justification: The paper includes a Limitations and Broader Impact section that discusses reward misspecification, synthetic speech bias, limited expert-evaluation scale, and misuse risks.

\item {\bf Theory assumptions and proofs}
    \item[] Question: For each theoretical result, does the paper provide the full set of assumptions and a complete (and correct) proof?
    \item[] Answer: \answerNA{} 
    \item[] Justification: The paper proposes an optimization objective but does not claim a formal theorem or proof.

    \item {\bf Experimental result reproducibility}
    \item[] Question: Does the paper fully disclose all the information needed to reproduce the main experimental results of the paper to the extent that it affects the main claims and/or conclusions of the paper (regardless of whether the code and data are provided or not)?
    \item[] Answer: \answerYes{} 
    \item[] Justification: The paper reports the training data mixture, speech-view construction, reward components, optimization hyperparameters, optimizer, baselines, datasets, metrics, judge protocol, uncertainty estimation, and compute resources needed to reproduce the main experiments.

\item {\bf Open access to data and code}
    \item[] Question: Does the paper provide open access to the data and code, with sufficient instructions to faithfully reproduce the main experimental results, as described in supplemental material?
    \item[] Answer: \answerNo{} 
    \item[] Justification: The submission does not include public code, generated data, checkpoints, or scripts.

\item {\bf Experimental setting/details}
    \item[] Question: Does the paper specify all the training and test details (e.g., data splits, hyperparameters, how they were chosen, type of optimizer) necessary to understand the results?
    \item[] Answer: \answerYes{} 
    \item[] Justification: The paper lists data splits, metrics, backbones, reward components, speech-view construction, and core hyperparameters.

\item {\bf Experiment statistical significance}
    \item[] Question: Does the paper report error bars suitably and correctly defined or other appropriate information about the statistical significance of the experiments?
    \item[] Answer: \answerYes{} 
    \item[] Justification: The experimental setting and appendix describe paired bootstrap confidence intervals for LC win rates, accuracy, consistency, and translation metrics. The paper does not claim repeated-training-run significance and lists larger repeated runs as a limitation.

\item {\bf Experiments compute resources}
    \item[] Question: For each experiment, does the paper provide sufficient information on the computer resources (type of compute workers, memory, time of execution) needed to reproduce the experiments?
    \item[] Answer: \answerYes{} 
    \item[] Justification: The appendix reports the GPU types used for teacher generation, speech synthesis, SFT, RL, and reward-weight validation.
    
\item {\bf Code of ethics}
    \item[] Question: Does the research conducted in the paper conform, in every respect, with the NeurIPS Code of Ethics \url{https://neurips.cc/public/EthicsGuidelines}?
    \item[] Answer: \answerYes{} 
    \item[] Justification: The submission is written to conform to the NeurIPS Code of Ethics and discusses anonymity, synthetic speech, evaluation bias, and potential misuse.

\item {\bf Broader impacts}
    \item[] Question: Does the paper discuss both potential positive societal impacts and negative societal impacts of the work performed?
    \item[] Answer: \answerYes{} 
    \item[] Justification: The Limitations and Broader Impact section discusses accessibility benefits as well as misuse risks such as voice-based misinformation and automated persuasion.
    
\item {\bf Safeguards}
    \item[] Question: Does the paper describe safeguards that have been put in place for responsible release of data or models that have a high risk for misuse (e.g., pre-trained language models, image generators, or scraped datasets)?
    \item[] Answer: \answerNA{} 
    \item[] Justification: The anonymous submission does not release checkpoints, generated speech data, or benchmark derivatives. The appendix specifies safeguards expected for any future release, including data provenance, model-card style restrictions, and documentation.

\item {\bf Licenses for existing assets}
    \item[] Question: Are the creators or original owners of assets (e.g., code, data, models), used in the paper, properly credited and are the license and terms of use explicitly mentioned and properly respected?
    \item[] Answer: \answerYes{} 
    \item[] Justification: Existing models, datasets, and speech assets are cited, and the appendix states that they are used under their original public licenses and terms without redistributing third-party assets in the anonymous submission. A full asset-license table should accompany any public artifact release.

\item {\bf New assets}
    \item[] Question: Are new assets introduced in the paper well documented and is the documentation provided alongside the assets?
    \item[] Answer: \answerNA{} 
    \item[] Justification: The paper creates generated speech views for training but does not release a new dataset, checkpoint, or code asset with the anonymous submission. The recipe, data mixture, and release expectations are documented for reproducibility.

\item {\bf Crowdsourcing and research with human subjects}
    \item[] Question: For crowdsourcing experiments and research with human subjects, does the paper include the full text of instructions given to participants and screenshots, if applicable, as well as details about compensation (if any)? 
    \item[] Answer: \answerYes{} 
    \item[] Justification: The expert reward audit uses qualified annotators to compare anonymized model outputs. The appendix provides the evaluation instruction, anonymization protocol, and compensation/role information; screenshots are not applicable because the audit uses text-form response pairs rather than an interactive interface.

\item {\bf Institutional review board (IRB) approvals or equivalent for research with human subjects}
    \item[] Question: Does the paper describe potential risks incurred by study participants, whether such risks were disclosed to the subjects, and whether Institutional Review Board (IRB) approvals (or an equivalent approval/review based on the requirements of your country or institution) were obtained?
    \item[] Answer: \answerNA{} 
    \item[] Justification: The expert audit evaluates anonymized model outputs rather than studying annotators, collects no personal data, and introduces no participant-facing intervention; IRB-style human-subject review is therefore not applicable.

\item {\bf Declaration of LLM usage}
    \item[] Question: Does the paper describe the usage of LLMs if it is an important, original, or non-standard component of the core methods in this research? Note that if the LLM is used only for writing, editing, or formatting purposes and does \emph{not} impact the core methodology, scientific rigor, or originality of the research, declaration is not required.
    \item[] Answer: \answerYes{} 
    \item[] Justification: The paper describes teacher LLMs, judge models, and SpeechLM backbones as core components of the proposed method.

\end{enumerate}

\end{document}